%% file: conditioned_attfuse_method.tex
\documentclass[10pt,twocolumn,letterpaper]{article}
\usepackage[pagenumbers]{cvpr}
\usepackage{microtype}
\definecolor{cvprblue}{rgb}{0.21,0.49,0.74}
\usepackage[pagebackref,breaklinks,colorlinks,allcolors=cvprblue]{hyperref}

\newcommand{\method}{\textsc{CoReQ}}

\def\confName{CVPR}
\def\confYear{2027}

\title{CoReQ: Condition-Aware Cross-Modal Repair and\\
Quality-Guided Robust Collaborative Perception}
\author{
Ziwei Guo$^{1}$ \quad
Lingzhi Li$^{2}$ \quad
Jin Wang$^{2}$ \quad
Yunjiang Xu$^{3}$\\
Yupeng Ouyang$^{1}$ \quad
Benyuan Yang$^{2}$ \quad
Yujun Shi$^{1}$\\[3pt]
$^{1}$School of Computer Science \& Technology, Soochow University\\
$^{2}$School of Future Science and Engineering, Soochow University\\
$^{3}$City University of Hong Kong\\
}
\date{}

\begin{document}
\maketitle

\begin{abstract}
Multi-vehicle collaborative perception can use observations from different perspectives to alleviate the problems of occlusion and limited sensing range. However, missing echoes and false echoes caused by bad weather will reduce the quality of single-vehicle features and affect collaborative detection. LiDAR provides richer geometric information, and 4D Radar has strong weather adaptability. The complementarity of the two provides the possibility to improve perception under degraded observations. However, how to adjust cross-modal repair according to the current observation conditions and how to reasonably integrate multi-vehicle features whose reliability still differs after repair are still problems that need to be solved. To this end, we propose CoReQ, a condition-aware repair and quality-guided collaborative perception framework. For the former, the condition-guided bidirectional cross-modal repair module (CBCR) extracts vehicle-level observation conditions from point cloud occupancy, density and feature statistics, and adjusts the bidirectional correction of LiDAR and Radar features accordingly. For the latter, we combine local reliability assessment with collaborative weighting: first, the task-aware quality estimation module (TaQE) uses the classification and positioning information of the EMA teacher to construct a supervision target and predict the spatial quality of the repaired features; then, the condition and quality guided fusion module (CoQF) matches the quality information with the content, and the observation conditions of both parties are jointly used for multi-vehicle feature weighted fusion. Experiments on V2X-R and V2X-Radar-C show that the framework is competitive under a variety of observation conditions, achieving significant gains in AP@0.7 on V2X-R under snowy conditions, and achieving the highest AP among the compared V2X-Radar-C methods.
\end{abstract}

\section{Introduction}
\label{sec:introduction}

\begin{figure}[!htbp]
  \centering
  \includegraphics[width=\columnwidth]{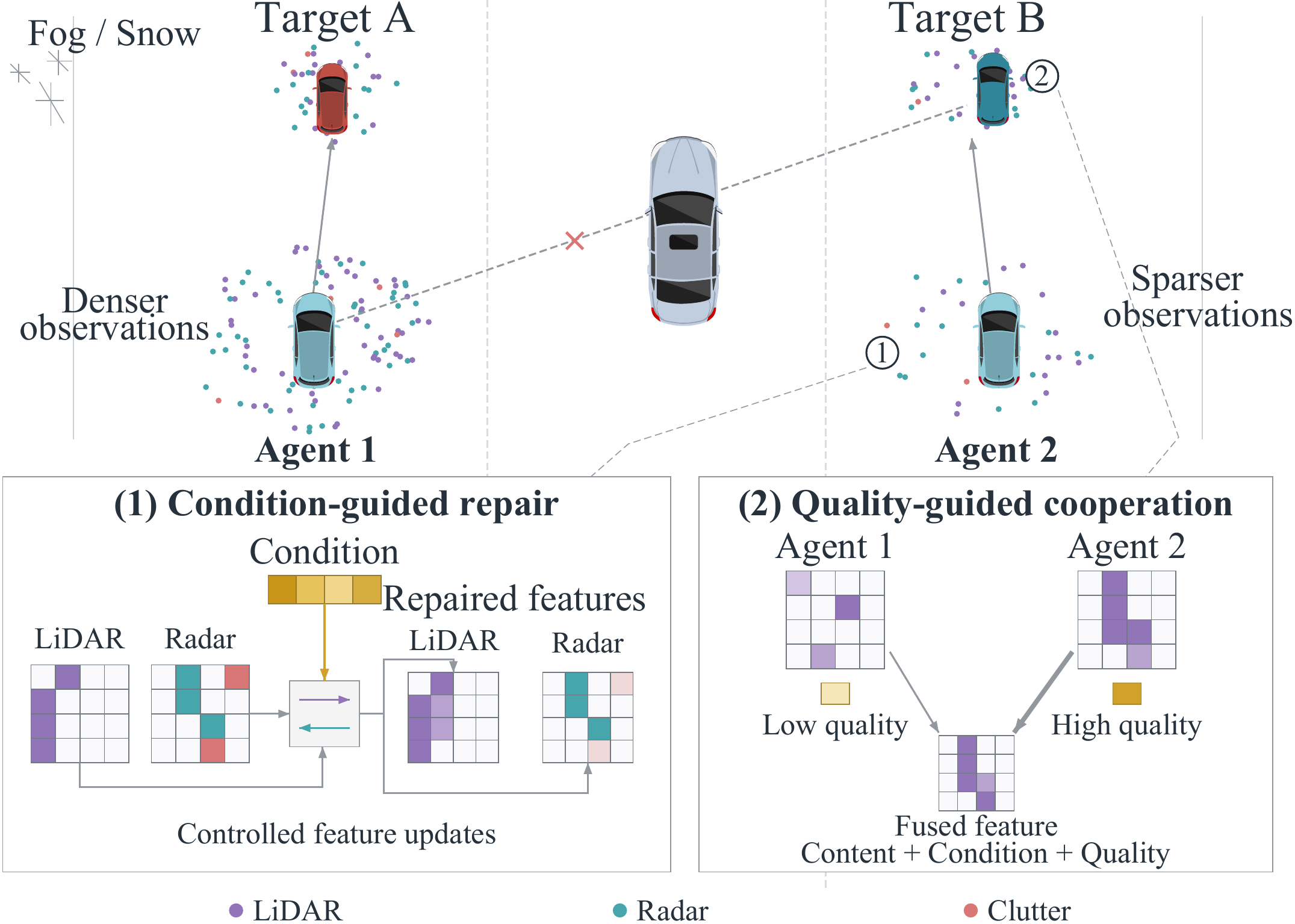}
  \caption{Condition-guided repair and quality-guided collaboration.
(1) Observation conditions guide bidirectional LiDAR--radar feature repair.
(2) Local task quality informs collaborative weighting, allowing an agent
with sparse overall observations to provide valuable local evidence.}
  \label{fig:motivation}
\end{figure}

Multi-agent collaborative perception expands the sensing range of individual vehicles and mitigates information loss caused by occlusions by sharing observations from different viewpoints. LiDAR, which provides rich geometric information, serves as a crucial input for 3D collaborative detection. However, in foggy or snowy weather, signal attenuation and backscattering from particles can lead to the loss of valid returns or the generation of false returns, thereby compromising the geometric representation of objects~\cite{hahner2021fog,hahner2022snow,rasshofer2011weather}. In this scenario, a collaborative system not only needs to acquire more observations but also must control the impact of degraded information on the fusion result.

Leveraging the complementarity between LiDAR and 4D radar is a key approach to improving perception under degraded conditions. While 4D radar offers robust performance across various weather conditions, its point clouds are relatively sparse and provide limited geometric detail. Single-vehicle methods, such as L4DR, REL, and RAF, enhance detection capabilities through techniques like multi-modal encoding, geometric augmentation, or reliability-aware fusion~\cite{huang2025l4dr,tong2026rel,park2026raf}. MDD and HRCP further introduce cross-modal processing into collaborative perception \cite{huang2025v2xr,yang2026hrcp}. Building on these studies, we further examine the interplay among observational conditions, cross-modal repair, and collaborative fusion: the global observational state of a vehicle is used to guide local repair, and the subsequent local detection quality, derived from the repaired features, is then employed to adjust the fusion of information across multiple vehicles.

This design is motivated by two considerations. First, observation coverage, sampling density, and feature distributions vary across different agents; consequently, similar local responses may correspond to different underlying observation contexts. For instance, a local gap in LiDAR data could represent a normal background feature or, alternatively, result from a failure to receive a return signal. Therefore, cross-modal repair requires integrating local features with the global observation context to dynamically adjust corrections derived from another modality, rather than treating any single modality as an invariably reliable reference.
Furthermore, the overall observation state cannot adequately characterize the quality of local detections. As illustrated in Figure~\ref{fig:motivation}, an agent with relatively sparse overall observations may still provide clear target information at specific locations, whereas an agent with denser overall observations might encounter local occlusions. While cross-modal repair can enhance feature representations, it does not guarantee that every location will support accurate classification and localization. Consequently, even after repair, it remains necessary to estimate spatial quality, which, alongside content matching and the respective observation conditions of the agents, should be incorporated into a collaborative weighting scheme.

Based on the analysis above, we propose CoReQ, which bridges condition-guided in-vehicle repair with quality-guided inter-vehicle collaboration. The Condition-Guided Bidirectional Cross-Modal Repair (CBCR) module extracts agent-level condition representations, derived from occupancy, density, and encoded features, to regulate constrained residual updates in both directions. The Task-Aware Quality Estimation (TaQE) module leverages local detection tasks and their corresponding EMA teacher models to assess the reliability of the repaired features in terms of both classification and localization. Finally, the Condition- and Quality-Guided Fusion (CoQF) module jointly models quality priors, pairwise conditional biases, and content similarity, enabling the adaptive aggregation of features from multiple vehicles across various BEV (Bird’s-Eye-View) locations.

The main contributions of this paper are as follows:
\begin{itemize}
    \item We propose \method, a LiDAR–4D radar collaborative detection framework designed for adverse weather conditions, which integrates vehicle-level observational contexts with local detection quality to guide cross-modal repair and multi-vehicle information fusion.

    \item We design a condition-guided bidirectional cross-modal repair mechanism, which integrates neighborhood observations and magnitude constraints to adjust feature updates. Furthermore, we employ an EMA-based teacher model to establish quality supervision that accounts for both classification and localization; the predicted quality, content matching, and the observation conditions of the ego and collaborating agents are then jointly utilized for collaborative weighting.

    \item Experimental validation on V2X-R and V2X-Radar-C demonstrates the effectiveness of this framework. Specifically, on the V2X-R dataset under snowy conditions, the AP@0.7 metric shows an improvement of 4.61 percentage points over the best-performing baseline; meanwhile, on V2X-Radar-C, the framework achieves the highest results across all three AP metrics among the methods compared. Ablation studies further analyze the contributions of the individual design components.
\end{itemize}

\begin{figure*}[!t]
    \centering
    \includegraphics[width=\textwidth]
    {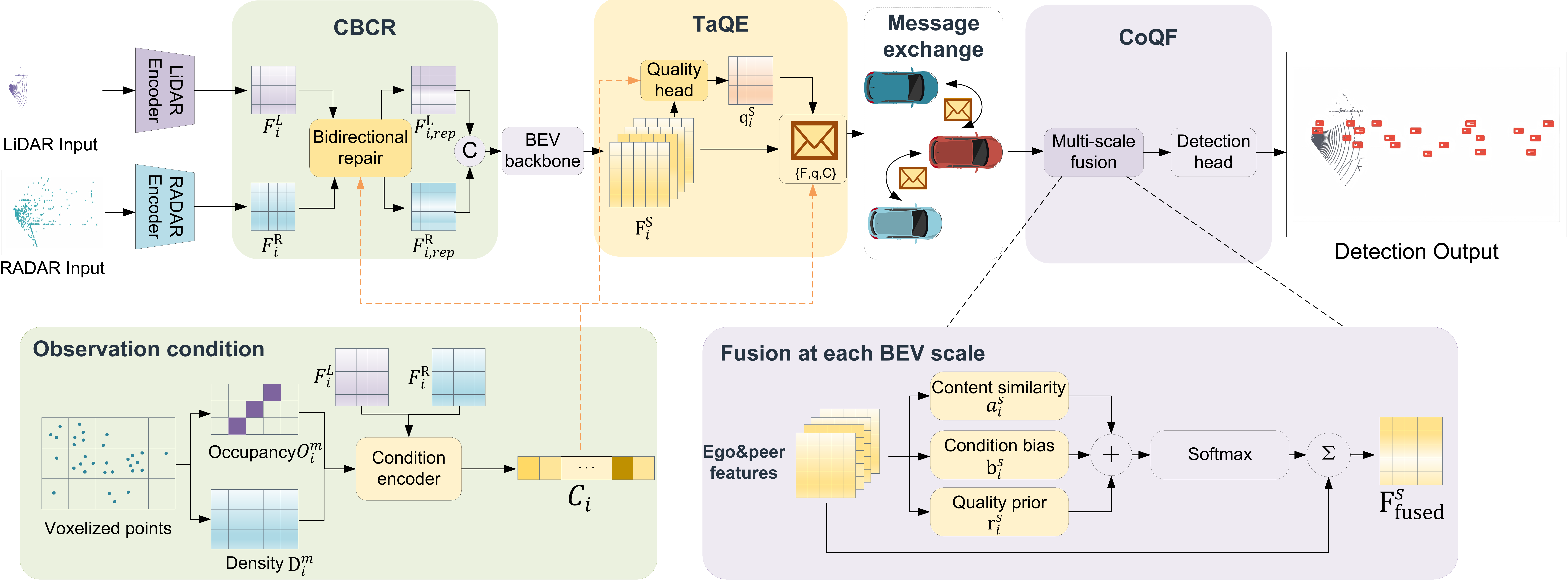}

    \caption{Overview of \method.
Observation conditions guide bidirectional LiDAR--radar repair.
The resulting BEV features, spatial quality maps, and condition
descriptors are used for collaborative fusion and final detection.}
    \label{fig:overview}
\end{figure*}
\section{Related Work}
\label{sec:related}

\paragraph{Multi-agent collaborative perception.}
Collaborative perception exchanges raw data, learned features, or detection results through early, intermediate, or late fusion. Intermediate fusion retains scene information while allowing flexible cross-agent interaction. Existing methods improve detection through feature aggregation, knowledge distillation, and attention~\cite{wang2020v2vnet,li2021disconet,xu2022v2xvit,xu2022opv2v}, or focus interaction on informative regions and instances~\cite{hu2022where2comm,xu2025instinct}.

To handle inconsistent observations, spatial calibration and temporal interaction improve alignment~\cite{lu2023coalign,zhang2024ermvp,yang2023how2comm,yang2023scope,chen2026catnet}, while reconstruction and distillation strengthen collaborative representations~\cite{wang2023core,zhang2025dsrc}. Multimodal collaboration further combines LiDAR and camera information within and across agents~\cite{xiang2023hmvit,zhao2023bm2cp}. While these advancements have improved information exchange among agents, adverse weather conditions can still degrade the quality of raw observations and the shared features extracted from them. Consequently, collaborative fusion must account for the reliability of these shared features in supporting accurate detection across different spatial locations.

\paragraph{Multimodal detection in adverse weather.}
Attenuation, scattering, and occlusion change point-cloud coverage and sampling distributions~\cite{hahner2021fog,hahner2022snow,rasshofer2011weather}. Corruption benchmarks document the resulting vulnerability of 3D perception systems~\cite{dong2023corruptions,kong2023robo3d}, while cross-weather distillation transfers knowledge from clear to degraded observations~\cite{huang2024sunshine}. Multimodal fusion and radar datasets support the use of complementary sensors under these conditions~\cite{bijelic2020fog,sheeny2021radiate,paek2022kradar}.

LiDAR–radar fusion improves detection under degraded observation conditions, through spatial feature interaction, multimodal denoising, and radar measurements~\cite{chae2024lrf,huang2025l4dr,chae2025doppler}, or radar-guided geometric supplementation~\cite{tong2026rel}. MDD and HRCP extend multimodal denoising or cross-modal enhancement to collaborative perception~\cite{huang2025v2xr,yang2026hrcp}. We focus on the connection between repair and fusion: observation conditions guide cross-modal updates, and the local detection quality of the repaired features informs multi-agent aggregation.

\section{Method}

\label{sec:method}

To improve multi-vehicle collaborative detection under adverse weather conditions, we propose CoReQ, which integrates observation-condition-guided cross-modal repair with local-quality-guided collaborative fusion. Section~\ref{sec:overview} presents the overall framework and information flow; Section~\ref{sec:repair} introduces the Condition-Guided Bidirectional Cross-Modal Repair (CBCR) module, which explains how the information exchange between LiDAR and Radar is adjusted based on current observation conditions; Section~\ref{sec:quality} describes the Task-Aware Quality Estimation (TaQE) module, which evaluates the local detection reliability of the repaired features; and Section~\ref{sec:fusion} details the Condition- and Quality-Guided Fusion (CoQF) module, which explains how this information is incorporated to perform multi-vehicle feature fusion.

\subsection{Overall Architecture}
\label{sec:overview}
As shown in Fig.~\ref{fig:overview}, for each vehicle $i$, the LiDAR and 4D radar point clouds are first encoded to extract features $F_i^L$ and $F_i^R$. Simultaneously, occupancy maps $O_i^m$ and density maps $D_i^m$ are computed for each modality $m \in \{L, R\}$. The CBCR module integrates these observations and encoded features to generate a condition vector $C_i$, which guides a bidirectional repair process for both modalities, yielding repaired features $F_{i,\mathrm{rep}}^L$ and $F_{i,\mathrm{rep}}^R$. These repaired features are concatenated and processed by a BEV backbone network to produce multi-scale representations $F_i^s$. Subsequently, the TaQE module utilizes $C_i$ to predict spatial quality maps $q_i^s$ for each scale $s$.

Neighboring vehicles transmit their multi-scale features, quality maps, and condition vectors to the ego-vehicle. The ego-vehicle aligns the received features and quality maps into its own coordinate system. Then, the CoQF module performs position-wise fusion by jointly considering content matching, observational conditions from both parties, and local quality, resulting in fused features $F_{\mathrm{fused}}^s$. Finally, a detection head outputs the object categories and 3D bounding boxes, denoted as $\widehat{\mathcal Y}_e$.

\subsection{Condition-Guided Bidirectional Cross-Modal Repair}
\label{sec:repair}
LiDAR and Radar provide different types of useful information depending on the observation conditions; therefore, the cross-modal repair process must be adjusted based on the current observational context. CBCR first extracts vehicle-level conditional representations, then integrates local features with neighborhood observations to govern feature repair in both directions.

\paragraph{Observation condition encoding.}
It is difficult to determine how to utilize information from another modality based solely on local responses. For instance, a local void could correspond to a normal background or, alternatively, result from a lack of return signals. To provide a more robust basis for such determinations, we construct a vehicle-level conditional representation by integrating point cloud coverage, sampling density, and learned feature statistics, as illustrated in Figure~\ref{fig:overview}.
Let $n_i^m(u)$ be the number of points, after voxelization, that remain within the pillar corresponding to spatial location $u$, and let $N_{\max}^m$ be the upper limit on the number of points per pillar for modality $m$. The occupancy map $O_i^m$ and the density map $D_i^m$ are defined as follows:
\begin{equation}
O_i^m(u)=\mathbf{1}[n_i^m(u)>0].
\label{eq:occupancy}
\end{equation}

\begin{equation}
D_i^m(u)=\min\left\{\frac{n_i^m(u)}{N_{\max}^m},1\right\}.
\label{eq:density}
\end{equation}
Occupancy and density describe the extent of observational coverage and sampling; however, since clutter can also inflate the number of points, we further incorporate learned feature responses. By applying mean and max pooling to occupied locations, where the mean response captures aggregate characteristics and the max response preserves salient information that might otherwise be diminished by averaging, we derive a feature summary, $V_i^m$. This summary, along with the occupancy rate $r_i^m$ and the mean normalized density of occupied grid cells $d_i^m$, is encoded into a condition vector $C_i \in \mathbb{R}^{16}$. This vector is learned through the detection task, requiring no weather-related labels, and serves to modulate the manner and intensity of cross-modal interactions.

\begin{figure}[!htbp]
  \centering
  \includegraphics[width=\columnwidth]
  {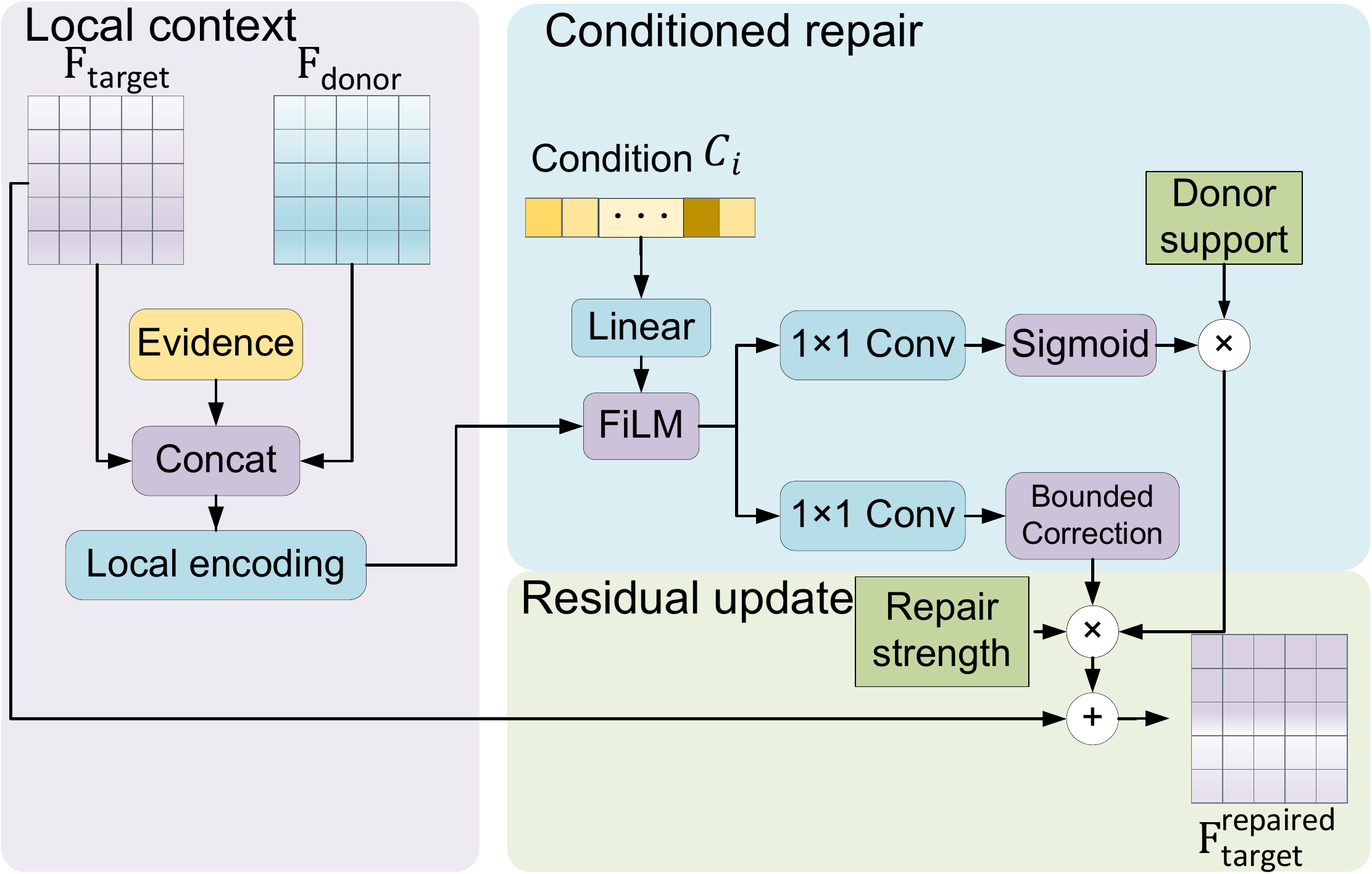}
  \caption{Condition-guided cross-modal repair.
The observation condition modulates a donor-to-target repair branch,
which applies support-gated, bounded residual corrections.}
  \label{fig:directional_repair}
\end{figure}
\paragraph{Condition-Guided Bidirectional Repair.}

The overall conditions dictate how the repair should align with the current observational state, while specific corrections require localized evidence. In any given repair direction, we designate the modality providing supplementary information as the donor modality and the modality undergoing repair as the target modality; these roles are reversed when the repair direction is inverted. Since echoes from the same target captured by two different sensors may fall into adjacent grid cells rather than identical ones, we perform a neighborhood expansion on the occupancy map to generate a support map, $S_i^m$, which helps determine whether the other modality has observations near the location being repaired. A repair gating mechanism integrates this support information with local features to decide which locations can be updated and to what extent. This approach enables the use of neighboring observations for supplementation while minimizing corrections that lack empirical observational support.
As shown in Fig.~\ref{fig:directional_repair}, we concatenate the target feature $F_{\mathrm{target}}$, the donor feature $F_{\mathrm{donor}}$, and occupancy and density evidence, and perform local encoding; the channel responses are then modulated by the condition $C_i$ through FiLM \cite{perez2018film}. Subsequently, the upper branch generates a spatial gate via convolution and sigmoid, and constrains the update locations with donor support; the lower branch predicts a bounded correction, which is used to enhance or suppress the target feature responses. The gate and the correction are multiplied element-wise, scaled by the repair strength, and added back to the original target feature to obtain $F_{\mathrm{target}}^{\mathrm{repaired}}$. The two directions use independent parameters and read the same set of original features. The repaired features of the two modalities are concatenated and encoded by a BEV backbone to form multi-scale representations.

\subsection{Task-Aware Quality Estimation}
\label{sec:quality}
\begin{figure}[!htbp]
  \centering
  \includegraphics[width=\columnwidth]{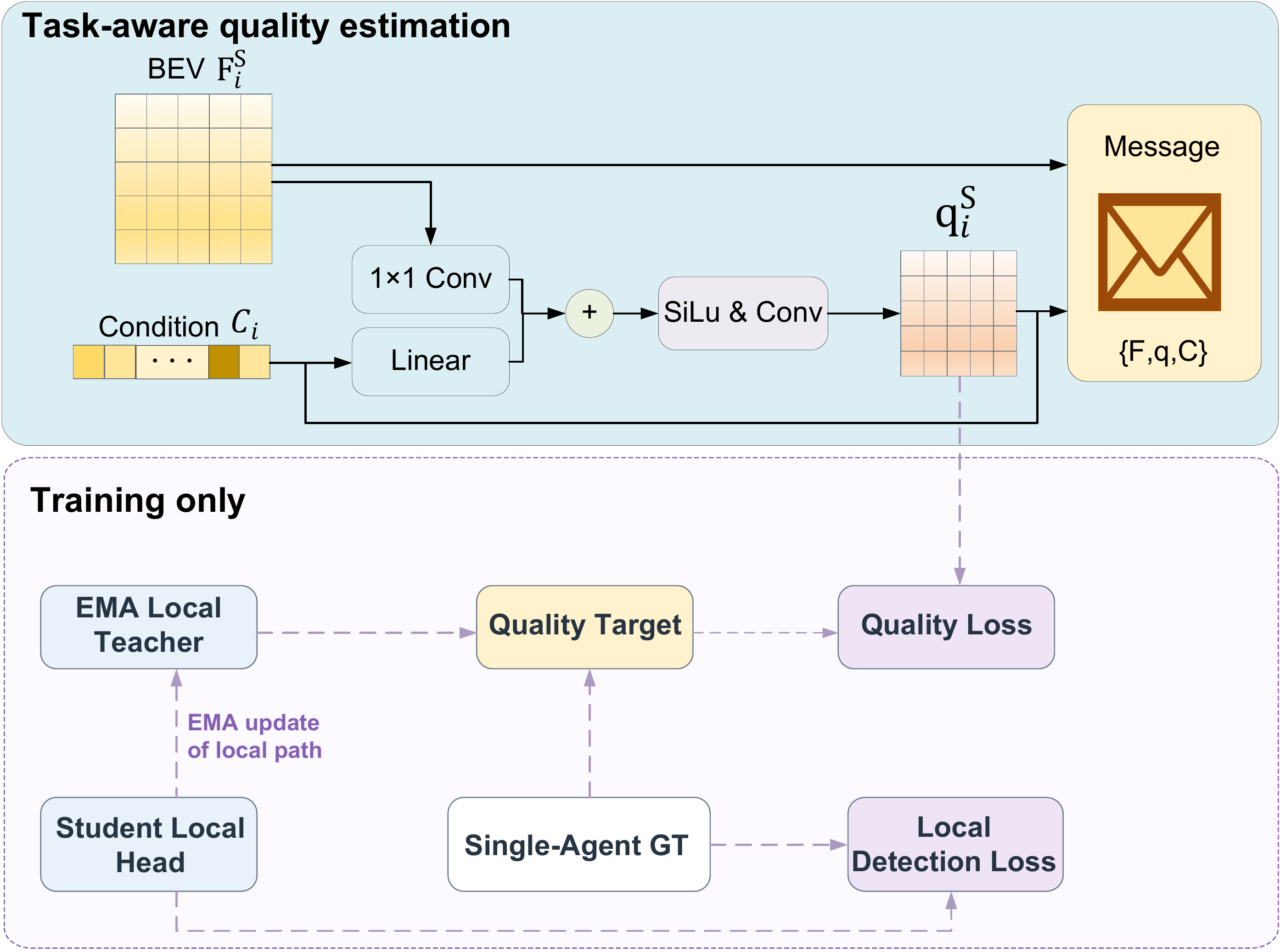}
  \caption{Task-quality estimation and supervision.
Spatial quality is predicted from BEV features and observation
conditions, with supervision provided by an EMA local teacher.}
  \label{fig:quality_architecture}
\end{figure}

While repair can improve feature representation, it cannot eliminate reliability disparities across all spatial locations. Factors such as occlusion, distance, and missing returns can lead to varying observation quality for the same vehicle across different targets, variations that cannot be adequately captured by vehicle-level conditions alone. Consequently, TaQE predicts spatial quality maps at multiple scales, evaluating the capacity of local features to support accurate classification and localization. This provides a basis for the subsequent collaborative selection of information, as illustrated in Figure~\ref{fig:quality_architecture}.

\paragraph{Condition-assisted quality prediction.}
Since the same local response can carry different meanings depending on the observational context, our quality prediction leverages both local BEV features and global conditions. We project the local BEV features and global conditions, sum them, and then apply a non-linear mapping followed by a sigmoid function to derive quality maps $q_i^s$ at various scales. The local features provide position-specific information, while the conditions supply the necessary background for their interpretation.
\paragraph{Task-specific quality supervision.}
Inspired by existing research on detection quality and spatial reliability \cite{jiang2018iou,li2020gfl,zhang2021varifocal,park2026raf}, we leverage local detection performance to supervise the spatial quality of the repaired features. An auxiliary local detector learns classification and localization tasks using single-agent annotations; its teacher model is updated via an EMA of the student model's parameters to provide relatively stable predictions \cite{tarvainen2017mean}. Since classification confidence does not adequately reflect localization accuracy, we construct a quality supervision objective by combining the teacher’s predicted foreground confidence with the IoU between the predicted bounding box and the ground-truth box.

\begin{equation}
y_{ia}=\bar p_{ia}\operatorname{IoU}_{\mathrm{BEV}}
(\bar b_{ia},b_{ia}^{*}).
\label{eq:quality_target}
\end{equation}
 $\bar p_{ia}$, $\bar b_{ia}$, and $b_{ia}^{*}$ represent the teacher’s foreground confidence, the predicted bounding box, and the matched ground-truth box for a positive anchor $a$, respectively. 

The quality map for the prediction is transmitted alongside the features and is used at the receiving end to calculate the weights for collaborative fusion.

\subsection{Condition and Quality Guided Fusion}
\label{sec:fusion}
Once the reconstructed features and quality maps for each vehicle are obtained, a decision must be made on how to combine this information at each specific location. Content matching provides a basis for local association, quality estimation reflects the reliability of individual vehicle detections, and the respective conditions offer complementary observational context. CoQF integrates these three elements to perform location-wise weighting, ensuring that the fusion process accounts for content alignment, detection reliability, and the underlying observational context.

\paragraph{Content matching.}
Using the ego vehicle as a reference, we first align the features of all vehicles to the ego vehicle coordinate frame, and then compute the feature similarity between neighboring vehicles and the ego vehicle at corresponding spatial locations; this serves as the basis for a collaborative scoring mechanism. Within a unified ego-centric coordinate system, we compute a content score via a scaled dot product:
\begin{equation}
a_i^s(u)=
\frac{\langle F_e^s(u),F_i^s(u)\rangle}{\sqrt{D_s}}.
\label{eq:content_similarity}
\end{equation}
Here, $D_s$ denotes the number of feature channels at scale $s$, and $a_i^s(u)$ represents the content similarity between vehicle $i$ and the ego-vehicle at position $u$ at scale $s$.
\paragraph{Condition-based adjustment.}
Similar local features may arise from different observational states; content matching alone cannot adequately account for discrepancies in observational coverage and sampling between the ego vehicle and neighboring vehicles. We jointly encode the conditions of both the ego vehicle and its neighbors, and integrate this with the local features of the neighboring vehicles to generate position-dependent conditional biases.
\begin{equation}
v_i^s = h_s([C_e,C_i]).
\label{eq:pair_condition}
\end{equation}

\begin{equation}
b_i^s(u) = \kappa_s\tanh\!\left(
\frac{\langle \phi_s(F_i^s(u)),v_i^s\rangle}{\sqrt{d_s}}
\right).
\label{eq:condition_bias}
\end{equation}
Here, \([C_e, C_i]\) represents the concatenation of conditions for the ego vehicle and neighboring vehicles; \(h_s\) generates paired condition representations via a linear mapping followed by a \(\tanh\) activation; \(\phi_s\) projects local features into a common dimension \(d_s\); and \(\kappa_s\) controls the magnitude of the bias.

\paragraph{Quality-guided.}
Even when feature content matches, local detection results may still be unreliable. Therefore, we use TaQE predictions as an additional indicator of reliability, applying a bounded logarithmic bias to moderately reduce the scores of low-quality observations:
\begin{equation}
r_i^s(u)=\beta_s
\max\!\left\{\log q_i^s(u),-\tau\right\}.
\label{eq:quality_bias}
\end{equation}
Here, \(\tau\) limits the maximum suppression magnitude, while \(\beta_s\) is a bounded, learnable coefficient. Even low-quality observations can still be incorporated through a fusion process that combines content matching with conditions derived from both sides.
\begin{table*}[!t]
\centering\small
\caption{Collaborative detection on V2X-R. }
\label{tab:v2xr}
\setlength{\tabcolsep}{4pt}
\begin{tabular*}{\textwidth}{@{\extracolsep{\fill}}lccccccccc@{}}
\toprule
& \multicolumn{3}{c}{Clean} & \multicolumn{3}{c}{Fog} &
\multicolumn{3}{c}{Snow}\\
\cmidrule(lr){2-4}\cmidrule(lr){5-7}\cmidrule(lr){8-10}
Method & AP@0.3 & AP@0.5 & AP@0.7 & AP@0.3 & AP@0.5 & AP@0.7 & AP@0.3 & AP@0.5 & AP@0.7\\
\midrule
AttFuse & 91.71 & 90.26 & 81.93 & 85.87 & 81.71 & 64.95 & 82.28 & 79.46 & 64.33\\
CoAlign & 91.92 & 90.76 & 83.06 & 85.22 & 82.00 & 65.04 & 82.63 & 80.47 & 63.95\\
CoBEVT & 92.21 & 91.28 & 79.48 & 86.26 & 82.90 & 64.59 & 86.69 & \textbf{84.49} & 65.28\\
\midrule
MDD & 90.50 & 89.20 & 80.68 & 86.75 & 82.92 & 66.82 & 82.98 & 80.27 & 64.49\\
L4DR & 87.69 & 86.40 & 77.23 & 79.57 & 74.20 & 57.20 & 77.78 & 75.27 & 59.17\\
HRCP  & 92.71 & 91.57 & \textbf{84.88} & 86.12 & 83.04 & \textbf{68.17} & 81.03 & 79.17 & 65.19 \\
\midrule
CoReQ (ours) & \textbf{93.21} & \textbf{91.94} & 84.15 & \textbf{87.42} & \textbf{83.61} & 67.19 & \textbf{86.94}& \textbf{84.49} & \textbf{69.89}\\
\bottomrule
\end{tabular*}
\end{table*}
\paragraph{Joint fusion.}
Three individual scores collectively determine the relative weight of each vehicle at a given location; these weights are then used to aggregate the aligned features.
\begin{equation}
\ell_i^s(u)=a_i^s(u)+b_i^s(u)+r_i^s(u).
\label{eq:fusion_score}
\end{equation}

\begin{equation}
\alpha_i^s(u)=
\operatorname{softmax}_{i\in\mathcal A_e}
\!\left(\ell_i^s(u)\right).
\label{eq:fusion_weight}
\end{equation}
where $\ell_i^s(u)$ denotes the combined fusion score for agent $i$ at position $u$ and scale $s$, and $\mathcal{A}_e$ includes the ego agent and its participating neighbors. Softmax normalizes these scores across agents to obtain the fusion weights $\alpha_i^s(u)$. The fused multi-scale features are resized to a common resolution and concatenated along the channel dimension before a detection head predicts object categories and 3D bounding boxes. The losses for classification and bounding box regression are used to jointly optimize the collaborative fusion, feature extraction, and repair processes.

\input{experiments_table_flow_fragment}

\section{Conclusion}
\label{sec:conclusion}
This paper proposes CoReQ, which guides cross-modal repair through observation conditions and uses the estimated spatial quality of the repaired features to guide multi-vehicle collaboration. CBCR combines condition modulation, donor support, and bounded residual updates to repair LiDAR and Radar features; TaQE learns spatial quality from the classification and localization results of a local EMA teacher; and CoQF jointly uses content similarity, pairwise conditions, and quality priors for multi-scale fusion. Experiments and ablations on V2X-R and V2X-Radar-C support this design, with particularly notable gains on the evaluated strict localization metrics in snowy weather.

\section*{LLM Usage Statement}
OpenAI Codex was used to assist in refining the content of the paper. Furthermore, it provided support for code development and debugging, the creation of ablation studies and evaluation scripts, and the articulation of the methodology.

{\small
\bibliographystyle{ieeenat_fullname}
\bibliography{conditioned_attfuse_references}
}
\end{document}

%% file: experiments_table_flow_fragment.tex

\section{Experiments}

\label{sec:experiments}

\subsection{Experimental Setup}
\label{sec:exp_setup}
\paragraph{Datasets and evaluation.}
We utilize two publicly available benchmarks for multimodal collaborative perception: V2X-R \cite{huang2025v2xr} and V2X-Radar-C \cite{yang2025v2xradar}. V2X-R is built upon CARLA~\cite{dosovitskiy2017carla} and OpenCDA, providing LiDAR, 4D radar, and camera data under both clear-weather and simulated fog/snow conditions; V2X-Radar-C is a multimodal dataset designed for collaborative perception in real-world scenarios, providing camera, LiDAR, and 4D radar data. In this work, we employ only LiDAR and 4D radar data, and we report the Average Precision (AP) at BEV IoU thresholds of 0.3, 0.5, and 0.7 for each respective dataset split.

\paragraph{Compared methods.}
The methods we compare against include MDD \cite{huang2025v2xr}, HRCP \cite{yang2026hrcp}, AttFuse \cite{xu2022opv2v}, CoBEVT \cite{xu2022cobevt}, CoAlign \cite{lu2023coalign}, and L4DR \cite{huang2025l4dr}. For collaborative baselines originally designed for single-modality inputs, we extract LiDAR and Radar features using separate pillar encoders, concatenate them along the channel dimension, feed the result into the BEV backbone, and retain their respective inter-vehicle fusion mechanisms. For the single-vehicle multi-modal L4DR, we incorporate a self-attention-based inter-vehicle fusion module into its multi-scale BEV backbone. Regarding multi-modal collaborative methods, MDD utilizes the pre-trained AttFuse+MDD model officially released by V2X-R, while HRCP employs the gating and cross-modal enhancement modules provided by its authors. All methods are evaluated using the same data splits and evaluation protocols as those applied to our proposed method.

\subsection{Implementation Details}
\label{sec:implementation}
For each scenario, one vehicle is designated as the ego-vehicle; vehicles located beyond the communication radius do not participate in the collaboration. The communication radii for V2X-R and V2X-Radar-C are 70 m and 300 m, respectively. The horizontal point cloud processing ranges are defined as $x \in [-140.8, 140.8]$ m and $y \in [-40, 40]$ m for the former, and $x, y \in [-102.4, 102.4]$ m for the latter, with the field-of-view filtering specific to each data loader also applied. The model employs an encoder based on PointPillars \cite{lang2019pointpillars}, with a horizontal pillar resolution of $0.4 \times 0.4$ m.
The model is implemented in PyTorch~\cite{paszke2019pytorch} and trained on an NVIDIA A100 GPU. We use the Adam optimizer~\cite{kingma2015adam} with an initial learning rate of $2 \times 10^{-3}$ and momentum coefficients of $(0.9, 0.999)$. The EMA teacher and the auxiliary local detector are used exclusively during the training phase. In the pose noise experiment, independent zero-mean Gaussian perturbations are added to the 3D position and heading angle of the collaborative neighboring vehicles. The translation standard deviation  $\sigma_t$ and the heading standard deviation $\sigma_r$ take paired values from 0.1 to 0.5, while the remaining pose parameters and the ego-vehicle pose remain unchanged. Detailed architectural parameters, auxiliary loss configurations, and the training schedule are provided in the supplementary material.

\subsection{Quantitative Results}
\label{sec:quantitative}
\paragraph{V2X-R Detection Performance.}
\label{sec:main_results}
\label{sec:weather_results}
Table~\ref{tab:v2xr} compares the performance of our method and the comparison method under different weather conditions on the V2X-R dataset. In sunny weather, CoReQ achieves 93.21\% and 91.94\% in AP@0.3 and AP@0.5, respectively, improving over HRCP by 0.50 and 0.37 percentage points; its AP@0.7 is 84.15\%, slightly lower than HRCP's 84.88\%. In foggy weather, CoReQ also achieves the highest results on AP@0.3 and AP@0.5, improving over the best comparison result by 0.67 and 0.57 percentage points, respectively, but its AP@0.7 is 0.98 percentage points lower than HRCP. These results indicate that CoReQ maintains good detection performance under both conditions, but its strict localization accuracy does not yet lead comprehensively.
In snowy weather, CoReQ achieves 86.94\% in AP@0.3 and 84.49\% in AP@0.5, obtaining the highest and tied-highest results, respectively; its AP@0.7 reaches 69.89\%, improving over the second-best CoBEVT by 4.61 percentage points and over HRCP by 4.70 percentage points. With AP@0.5 tied with the best comparison method, the more pronounced advantage in AP@0.7 indicates that CoReQ's improvement in snowy weather is mainly reflected under stricter localization requirements.


\begin{figure}[!htbp]
\centering
\includegraphics[width=\columnwidth]{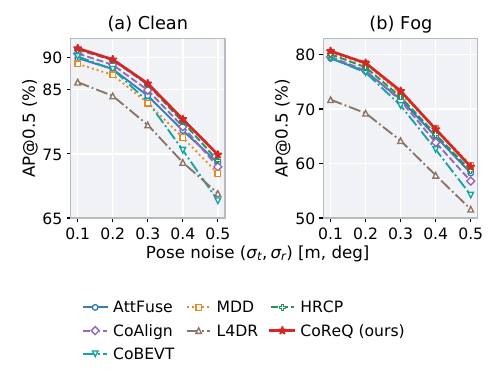}
\caption{Detection performance measured by AP@0.5 under pose noise on V2X-R.
}
\label{fig:pose_noise}
\end{figure}

\paragraph{Robustness to pose noise.}
\label{sec:pose_noise}
Figure~\ref{fig:pose_noise} shows the impact of registration error between test vehicles in sunny and foggy weather.
CoReQ achieves the highest AP@0.5 under all tested pose noise levels in clear weather. Under foggy conditions, it remains superior when the pose perturbation parameters $(\sigma_t,\sigma_r)$ increase from $(0.1,0.1)$ to $(0.3,0.3)$, but is slightly lower than MDD at $(0.4,0.4)$ and $(0.5,0.5)$. As the perturbation increases, the detection performance of all methods decreases, indicating that spatial misalignment still affects the utilization of collaborative information. CoReQ maintains a leading position under most settings, but its observation-condition and quality-guided mechanisms do not directly correct pose errors, and its advantage weakens somewhat under larger foggy-weather perturbations.


\paragraph{V2X-Radar-C Detection Performance.}
\label{sec:radarc_results}

\begin{table}[!htbp]
\centering\small
\caption{Collaborative detection on V2X-Radar-C. }
\label{tab:radarc}
\setlength{\tabcolsep}{3.5pt}
\begin{tabular*}{\columnwidth}{@{\extracolsep{\fill}}lccc@{}}
\toprule
Method & AP@0.3 & AP@0.5 & AP@0.7\\
\midrule
AttFuse & 76.53 & 70.31 & 51.01\\
CoAlign & 80.37 & 74.51 & 54.01\\
CoBEVT & 82.29 & 77.38 & 49.95\\
\midrule
MDD & 78.07 & 72.54 & 47.60\\
L4DR & 82.79 & 77.38 & 58.65\\
HRCP & 82.38 & 76.32 & 57.55\\
\midrule
CoReQ (ours) & \textbf{83.82} & \textbf{78.31} & \textbf{58.78}\\
\bottomrule
\end{tabular*}
\end{table}

Table~\ref{tab:radarc} presents the detection performance of all methods on the real-world dataset V2X-Radar-C. CoReQ achieves AP@0.3, AP@0.5, and AP@0.7 scores of 83.82\%, 78.31\%, and 58.78\%, respectively, outperforming the best existing baseline results by 1.03, 0.93, and 0.13 percentage points. Compared to AttFuse, these metrics show improvements of 7.29, 8.00, and 7.77 percentage points; relative to HRCP, the gains are 1.44, 1.99, and 1.23 percentage points; and compared to MDD, they are 5.75, 5.77, and 11.18 percentage points higher. Furthermore, for AP@0.7, CoReQ surpasses CoAlign and CoBEVT by 4.77 and 8.83 percentage points, respectively. In real-world scenarios, factors such as distance, occlusion, and viewpoint variation lead to inconsistent observation quality; conditional repair and local quality fusion address these challenges by focusing on the overall observation of vehicles and the reliability of local information, respectively. When compared to L4DR, CoReQ’s AP@0.7 score is only 0.13 percentage points higher, indicating that the performance gains over L4DR are primarily concentrated in the AP@0.3 and AP@0.5 metrics, while performance in terms of strict localization remains quite similar.

\begin{figure*}[!t]
  \centering
  \begingroup
  \setlength{\fboxrule}{0.4pt}
  \setlength{\fboxsep}{0pt}
  
  \newcommand{\qualcolwidth}{%
    \dimexpr.25\textwidth-3pt\relax
  }
  
  \setkeys{Gin}{
    width=\dimexpr.25\textwidth-3pt-2\fboxrule\relax,
    height=.16\textwidth
  }
  
  \newcommand{\qualpanel}[1]{%
    {\color{black}\fbox{#1}}%
  }
  
  \noindent
  \hspace*{12pt}%
  \makebox[\qualcolwidth]{\textbf{AttFuse}}%
  \makebox[\qualcolwidth]{\textbf{CoAlign}}%
  \makebox[\qualcolwidth]{\textbf{MDD}}%
  \makebox[\qualcolwidth]{\textbf{CoReQ (ours)}}%
  \par\vspace{2pt}
  
  \offinterlineskip
  \hbox{%
    \makebox[12pt]{\rotatebox{90}{\small Fog}}%
    \qualpanel{\includegraphics{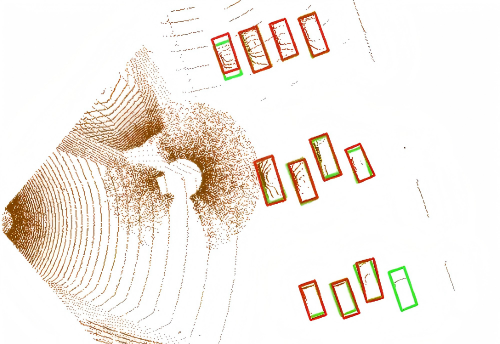}}%
    \qualpanel{\includegraphics{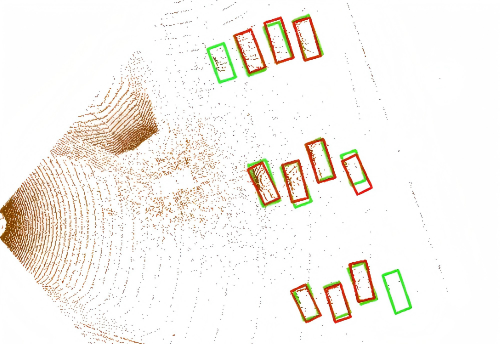}}%
    \qualpanel{\includegraphics{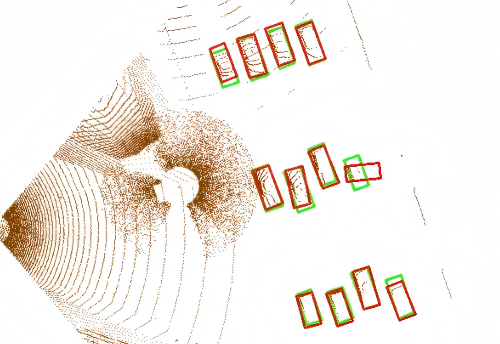}}%
    \qualpanel{\includegraphics{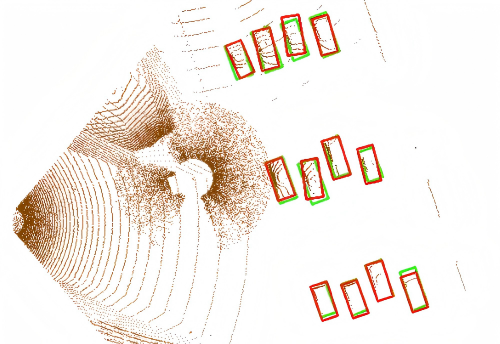}}%
  }
  \hbox{%
    \makebox[12pt]{\rotatebox{90}{\small Snow}}%
    \qualpanel{\includegraphics{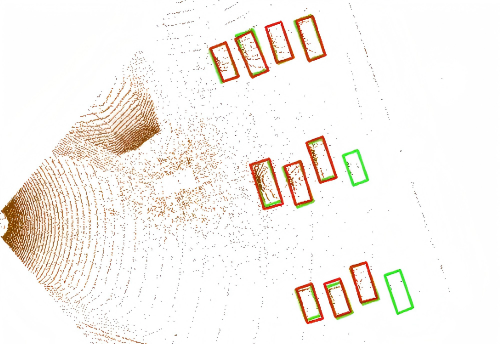}}%
    \qualpanel{\includegraphics{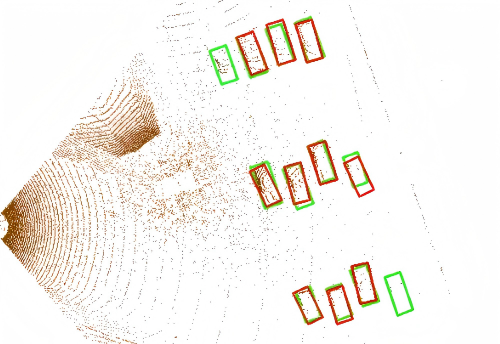}}%
    \qualpanel{\includegraphics{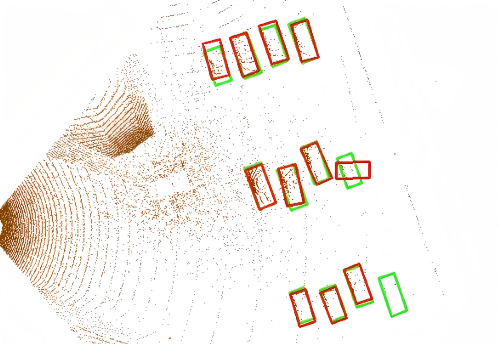}}%
    \qualpanel{\includegraphics{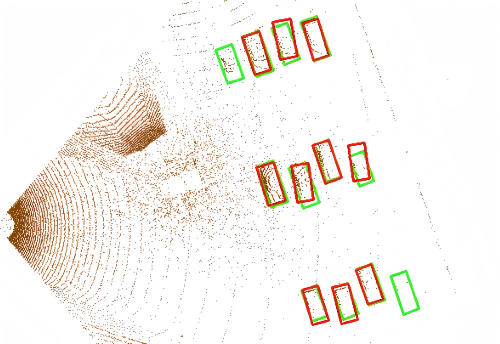}}%
  }
  \endgroup
  
  \caption{The visualization of detection results of AttFuse, CoAlign, MDD, and Ours under foggy and snowy conditions in V2X-R.
Green boxes represent ground-truth bounding boxes,
and red boxes represent predicted bounding boxes.}
  \label{fig:qualitative_weather}
  \end{figure*}

\subsection{Ablation Studies}
\label{sec:ablations}
\paragraph{Component Ablation.}
Table~\ref{tab:components} compares four cumulative configurations for V2X-R under foggy and snowy weather conditions. After adding CBCR, AP@0.5 in foggy and snowy weather improves by 0.90 and 0.65 percentage points, respectively, and AP@0.7 also improves in both cases, indicating that condition-guided cross-modal repair helps improve detection performance. After further adding CF, the metrics under both weather conditions continue to improve, with AP@0.7 increasing by 2.97 and 2.95 percentage points, respectively, showing that fusing with both parties' observation conditions also helps strict localization.
After adding QEF, AP@0.5 and AP@0.7 in snowy weather improve by 3.23 and 2.45 percentage points, respectively. The combination of local detection supervision and spatial quality weighting provides support for handling reliability differences at different locations in snowy weather. In foggy weather, AP@0.5 improves by 0.59 percentage points, but AP@0.7 decreases by 0.48 percentage points, indicating that the gains of this combination do not cover all metrics. Compared with the matched baseline, the full model achieves cumulative improvements of 2.94 and 3.93 percentage points in AP@0.5 and AP@0.7 in foggy weather, and 5.07 and 5.97 percentage points in snowy weather, respectively.

\begin{table}[!htbp]
\centering
\caption{Progressive component ablation on V2X-R.
CF: condition-guided fusion;
QEF: quality estimation and fusion.
}
\label{tab:components}
\footnotesize
\setlength{\tabcolsep}{2pt}
\begin{tabular*}{\columnwidth}{
@{\extracolsep{\fill}}ccccccc@{}}
\toprule
\multicolumn{3}{c}{Components}
& \multicolumn{2}{c}{Fog}
& \multicolumn{2}{c}{Snow}\\
\cmidrule(lr){1-3}
\cmidrule(lr){4-5}
\cmidrule(lr){6-7}
CBCR & CF & QEF
& AP@${0.5}$ & AP@${0.7}$
& AP@${0.5}$ & AP@${0.7}$\\
\midrule
& & 
& 80.67 & 63.26
& 79.42 & 63.92\\

$\checkmark$ & &
& 81.57 & 64.70
& 80.07 & 64.49\\

$\checkmark$ & $\checkmark$ &
& 83.02 & \textbf{67.67}
& 81.26 & 67.44\\

$\checkmark$ & $\checkmark$ & $\checkmark$
& \textbf{83.61} & 67.19
& \textbf{84.49} & \textbf{69.89}\\
\bottomrule
\end{tabular*}
\end{table}

\paragraph{Input modalities.}
Table~\ref{tab:modalities} presents the modality ablation results on V2X-R. In both foggy and snowy weather conditions, the fused input outperforms single-modality configurations in terms of AP@0.5 and AP@0.7. Specifically, the fused approach yields improvements of 10.86 and 17.65 percentage points in AP@0.7 compared to LiDAR-only, and 18.78 and 22.36 percentage points compared to Radar-only. In the single-mode results, Radar-only had a higher AP@0.5, while LiDAR-only had a better AP@0.7, reflecting the weather adaptability of Radar and the geometric representation advantage of LiDAR. Fusing these two modalities enables both reliable object detection and accurate localization under degraded environmental conditions.

\begin{table}[!htbp]
\centering\small
\caption{Modality analysis on V2X-R under fog and snow.}
\label{tab:modalities}
\footnotesize
\setlength{\tabcolsep}{2pt}
\begin{tabular*}{\columnwidth}{@{\extracolsep{\fill}}lcccc@{}}
\toprule
& \multicolumn{2}{c}{Fog} & \multicolumn{2}{c}{Snow}\\
\cmidrule(lr){2-3}\cmidrule(lr){4-5}
Input  & AP@${0.5}$ & AP@${0.7}$
& AP@${0.5}$ & AP@${0.7}$\\
\midrule
LiDAR only & 63.88 & 56.33 & 66.00 & 52.24\\
4D radar only & 74.77 & 48.41 & 73.48 & 47.53\\
LiDAR + 4D radar & \textbf{83.61} & \textbf{67.19} & \textbf{84.49} & \textbf{69.89}\\
\bottomrule
\end{tabular*}
\end{table}

\subsection{Qualitative Results}
\label{sec:qualitative}
Figure~\ref{fig:qualitative_weather} presents the detection results of AttFuse, CoAlign, MDD, and CoReQ in V2X-R scenarios under foggy and snowy conditions. Each row corresponds to the same frame, with consistent visualization settings applied across all methods. In foggy scenes, AttFuse and CoAlign missed some targets in the lower right corner, and MDD showed a significant deviation in predicting the orientation of some vehicles in the middle. In contrast, CoReQ detected these targets, and some of the predicted boxes matched the ground truth boxes in terms of position and orientation more closely. In the snowy scene, CoReQ maintains relatively strong localization performance for several vehicles in the central area, though some missed detections still occur. These examples provide an intuitive illustration of the differences among the various methods regarding object detection and localization, serving as a complement to the quantitative evaluations.